# Machine Learning Applications in Lung Cancer Diagnosis, Treatment and Prognosis

Yawei Li[1], Xin Wu[2], Ping Yang[3], Guoqian Jiang[4], Yuan Luo[1,*]

*[1] Department of Preventive Medicine, Northwestern University, Feinberg School of Medicine, Chicago, IL 60611, USA*

*[2] Department of Medicine, University of Illinois at Chicago, Chicago, IL 60612, USA*

*[3] Department of Quantitative Health Sciences, Mayo Clinic, Rochester, MN, 55905 Arizona, AZ, USA*

*[4] Department of Artificial Intelligence and Informatics, Mayo Clinic, Rochester, MN, 55905, USA*

[*] Corresponding author. Email: yuan.luo@northwestern.edu (Luo Y)

**Running title:** *Li et al / Machine Learning Applications in Lung Cancer*

## Abstract

The recent development of imaging and sequencing technologies enables systematic advances in the clinical study of lung cancer. Meanwhile, the human mind is limited in effectively handling and fully utilizing the accumulation of such enormous amounts of data. Machine learning-based approaches play a critical role in integrating and analyzing these large and complex datasets, which have extensively characterized lung cancer through the use of different perspectives from these accrued data. In this article, we provide an overview of machine learning-based approaches that strengthen the varying aspects of lung cancer diagnosis and therapy, including early detection, auxiliary diagnosis, prognosis prediction and immunotherapy practice. Moreover, we highlight the challenges and opportunities for future applications of machine learning in lung cancer.

## Introduction

Lung cancer is one of the most frequently diagnosed cancers and the leading cause of cancer deaths worldwide -- about 2.20 million new patients are diagnosed with lung cancer each year [1], and 75% of whom die within five years of diagnosis [2]. High intra-tumor heterogeneity (ITH) and complexity of cancer cells giving rise to drug resistance make cancer treatment more challenging [3]. Over the past decades, the continuous evolution of technologies in cancer research has contributed to many large collaborative cancer projects which have generated numerous clinical, medical imaging and sequencing databases [4-6]. These databases facilitate researchers in investigating comprehensive patterns of lung cancer from diagnosis, treatment and responses, to clinical outcomes [7]. In particular, current studies on -omics analysis, such as genomics, transcriptomics, proteomics, and metabolomics, have expanded our tools and capabilities for research. Cancer studies are undergoing a shift towards the integration of multiple data types and mega sizes. However, using diverse and high-dimensional data types for clinical tasks requires significant time and expertise even with assistance from dimension reduction methods such as matrix and tensor factorizations [8-11], and analyzing the exponentially growing cancer-associated databases poses a major challenge to researchers. Therefore, using machine learning (ML) models to automatically learn the internal characteristics of different data types to assist physicians' decision-making has become increasingly important.

ML is a subgroup of artificial intelligence (AI) that focuses on making predictions by identifying patterns in data using mathematical algorithms [12]. It has served as an assisting tool in cancer phenotyping and therapy for decades [13-19], and has been widely implemented in advanced approaches for early detection, cancer type classification, signature extraction, tumor microenvironment (TME) deconvolution, prognosis prediction, and drug response evaluation [20-27]. Herein we present an overview of the main ML algorithms that have been used to integrate complex biomedical data (including, e.g., imaging or sequencing data) for different aspects of lung cancer (**Figure 1**, Table S1 and S2), and outline major challenges and opportunities for future applications of ML in lung cancer clinical research and practice. We hope that this review promotes a better understanding of the roles and potentialities of ML in this field.

## Apply ML for early detection and auxiliary diagnosis of lung cancer

**ML on early detection and diagnosis using medical imaging datasets**

Early diagnosis is an important procedure for reducing deaths related to lung cancer. Chest screening using low-dose computed tomography (CT) is the primary approach for surveillance of people with increased lung cancer risk. To promote diagnostic efficiency, the computer-aided diagnosis (CAD) system was developed to assist physicians in the interpretation of medical imaging data [28, 29], which has been demonstrated as a useful second opinion for physicians [30]. The original CAD task can be broken into two steps: nodule feature extraction and clinical judgment inference (classification). Some approaches apply the measured texture features of specified nodules in CT images combined with the patient's clinical variables as input features to train a machine learning classifier, including logistic regression (LR) [31-33] or linear discriminant analysis (LDA) [34], for malignancy risk estimation. Typically, these measurements include nodule size, nodule type, nodule location, nodule count, nodule boundary and emphysema information in CT images, and the clinical variables include the patient's age, gender, specimen collection timing, family history of lung cancer, smoking exposure, and more. However, these features are mostly subjective and arbitrarily-defined, and usually fail to achieve a complete and quantitative description of malignant nodule appearances.

With the development of deep learning algorithms, especially convolutional neural networks (CNNs), more studies have been conducted to apply CNN-based models in the CAD system to improve its accuracy and reduce its false positive rate and execution time during lung tumor detection [35, 36]. The workflow of these models usually consists of three steps: nodule detection and segmentation, nodule feature extraction, and clinical judgment inference [37]. Unlike traditional CAD systems, the CNN-based CAD system can automatically retrieve and extract intrinsic features of a suspicious nodule [38, 39], and can model the 3D shape of a nodule. For example, Ciompi et al. [40] designed a model based on OverFeat [41, 42] by extracting three 2D-view-feature vectors (axial, coronal and sagittal) of the nodule from CT scans. The recently integrated CNN models facilitate a global and comprehensive inspection of nodules for feature characterization from CT images. Buty et al. [43] designed a complementary CNN model, where a spherical harmonic model [44] for nodule segmentation was used to obtain the shape descriptions ("shape" feature) of the segmented nodule and a deep convolutional neural networks (DCNN)-based model [41] to extract the texture and intensity features ("appearance" feature) of the nodule. The downstream classification relied on the combination of "shape" and "appearance" features. Similarly, Venkadesh et al. [45] used an ensemble model from two different models, 2D-ResNet50-based [46] and 3D-inception-V1 [47], to respectively extract two features of a pulmonary nodule, and then concatenated the two features as the input features for classification. A superiority of the ensemble CNN model is that, it can accurately detect different sizes of nodules with strong discriminative power using the raw CT images. Benefiting from the features extracted from state-of-the-art CNN models, clinical judgment inference can be implemented through frequent ML techniques, including LR, random forest (RF), support vector machine (SVM) and neural-networks (NNs). Notably, some studies also employed CNN models for final clinical judgment inference. Ardila et al. [48] proposed an end-to-end approach to systematically model both localization and lung cancer risk categorization tasks using the input CT data alone. Their approach was based on a combination of three CNN models: a Mask-RCNN [49] model for lung tissue segmentation, a modified RetinaNet [50] model for cancer region of interest (ROI) detection, and a full-volume model based on 3D-inflated inception-V1 [51, 52] for malignancy risk prediction.

In addition to CT images, CNN-based models are also widely used in histological imaging to help with lung cancer diagnosis. Compared with CT imaging, histological imaging can provide more biological information about cancer at the cellular level. To this end, AbdulJabbar et al. [53] used the Micro-Net [54] model to identify tissue boundaries followed by an SCCNN [55] model to segment individual cells from hematoxylin and eosin (H&E)-stained and immunohistochemistry (IHC) images. The segmented cells were then applied for cell type classification to evaluate the proportions of each cell type in the images. Another study [56] utilized the inception-V3 network [57] to classify whether the tissue was lung adenocarcinoma (LUAD), lung squamous cell carcinoma (LUSC), or normal from H&E-stained histopathology whole-slide images. It is worth noting that this model can also predict whether a given tissue has somatic mutations in several lung cancer driver genes, including *STK11*, *EGFR*, *FAT1*, *SETBP1*, *KRAS* and *TP53*.

**ML on early detection and diagnosis using -omics sequencing datasets**

Although periodic medical imaging tests are recommended for high-risk populations, implementation has been complicated by a high false-discovery rate [58, 59]. Therefore, there is a critical need for new techniques in early detection of lung cancers. Recent sequencing technologies enable diverse methods for early detection of lung cancer [60]. In the meantime, accurately classifying lung cancer subtypes is crucial in guiding optimal therapeutic decision-making. LUAD (~45%) and LUSC (~25%) are the two most common subtypes of lung cancer but often treated similarly except for targeted therapy [61]. However, studies have indicated that LUAD and LUSC have drastically different biological signatures, and they have suggested that LUAD and LUSC should be classified and treated as different cancers [62, 63]. From a computational perspective, both early detection and subtype identification are part of the classification task. Previous ML studies have shown the efficiency and advancement of early detection and cancer type classification in large pan-cancer sequencing datasets [64-72], which may provide evidence for lung cancer diagnosis. It is known that cancer cells are characterized by many genetic variations, and the accumulation of these genetic variations can be signatures that document the mutational patterns of different cancer types [3, 5, 73, 74]. For this reason, recent studies have concentrated on extracting better genomic signatures as input features to boost the accuracy of their ML models. For early detection, blood-based liquid biopsy, including cell-free DNA (cfDNA) fragments, circulating tumor DNA (ctDNA), microRNA (miRNA), methylation, exosomes and circulating tumor cells (CTCs), to explore potential circulating tumor signatures is considered a reliable method [60]. Integrating these liquid biopsy signatures, many discriminative models (SVM, RF, LR) have been used to detect tumors with high discovery rates [75-78]. For lung cancer subtype classification, somatic mutations, including single-nucleotide variants (SNVs), insertions, and deletions, usually have specific cancer type profiles [79]. Thus, studies have leveraged somatic mutations as input features to train classifiers for LUAD-LUSC classification [80]. Many of these mutations, especially driver mutations, can change expression levels, which impact gene function and interrupt cellular signaling processes [79]. As a result, different cancer types show different expression levels of certain proteins [81, 82]. Imposed by these unique cancer-type-expression-profiles, ML models can leverage RNA sequencing as input data to categorize the malignancy (benign or malignant) and subtypes (LUAD or LUSC) of patients [83-86]. Similarly, copy number variation (CNV) is reported to be highly correlated with differential gene expression [87], and can be ubiquitously detected in cancer cells. As such, CNVs can also be used to train ML models for cancer type classification in lung cancer studies [78, 88, 89]. More recently, Jurmeister et al. [90] used DNA methylation profiles as input features to determine if the detected malignant nodule is primary lung cancer or the metastasis of another cancer. Directly using all generated genes as an input feature may result in overfitting [91]. Thus, many studies used different computational approaches to select multiple cancer-associated genes to enhance their ML models. Some studies used ML based algorithms for feature selection. For example, Liang et al. [77] and Whitney et al. [83] employed the least absolute shrinkage and selection operator (LASSO) method to select the optimal markers for model training; Aliferis et al. [86] utilized recursive feature elimination (RFE) [92] and univariate association filtering (UAF) models to select highly cancer-associated genes. Apart from ML-based models, some studies used statistical methods for feature selection. Raman et al. [78] designed a copy number profile abnormality (CPA) score to reinforce the CNV feature which is more robust and less subject to variable sample quality than directly using CNVs as the input feature. Daemen et al. [89] integrated several statistical tests (ordinary fold changes, ordinary t-statistics, SAM-statistics and moderated t-statistics) to select a robust differential expression gene set. Aside from these single-measured signatures, some studies [78, 83, 85] combined the -omics signatures with clinical signatures to achieve better results. Using these tumor-type specific -omics signatures, many algorithms (K-Nearest Neighbors (KNN), naive Bayes (NB), SVM, decision tree (DT), LR, RF, LDA, gradient boosting and NN) have demonstrated their ability to accurately detect and classify different lung cancer patterns (**Table 1**). It is note that to improve the accuracy of ML models, Kobayashi et al. [80] added an element-wise input scaling for the neural network model, which allows the model to maintain its accuracy with a small number of learnable parameters for optimization.

## Apply ML for lung cancer treatment response and survival prediction

### Prognosis and drug response prediction

Sophisticated ML models have acted as supplements for cancer intervention response evaluation and prediction [93, 94], and have demonstrated advances in optimizing therapy decisions that improve chances of successful recovery [95, 96]. There are several metrics that are available for evaluating cancer therapy response, including the response evaluation criteria in solid tumors (RECIST) [97]. The definition of RECIST relies on imaging data, mainly CT and magnetic resonance imaging (MRI), to determine how tumors grow or shrink in patients [98]. To track the tumor volume changes from CT images, Jiang et al. [99] designed an integrated CNN model. Their CNN model used two deep networks based on a full-resolution residual networks [100] model by adding multiple residual streams of varying resolutions, so that they could simultaneously combine features at different resolutions for segmenting lung tumors. With the RECIST criterion, Qureshi [101] set up a molecular dynamics simulation with a machine learning model to predict the RECIST level under *EGFR* Tyrosine kinase inhibitors (TKIs) therapy given the patient's mutation profile in gene *EGFR*. In a recent study, the authors defined a different metric, tumor proportional scoring (TPS) calculated as the percentage of tumor cells in digital pathology images, to evaluate the lung cancer treatment response [102]. They applied the Otsu threshold [103] with an auxiliary classifier generative adversarial network (AC-GAN) model to identify positive tumor cell regions (TC(+)) and negative tumor cell regions (TC(−)). And ultimately used the ratio between the pixel count of the TC(+) regions and the pixel count of all detected tumor cell regions to evaluate the TPS number. Another study from Geeleher et al. [104] used half-maximal inhibitory concentration (IC50) to evaluate drug response. In their model, the authors applied a ridge regression model [105] to estimate IC50 values for different cell lines in terms of their whole-genome expression level.

**Survival prediction**

Prognosis and survival prediction as a part of clinical oncology is a tough but essential task for physicians, as knowing the survival period can inform treatment decisions and benefit patients in managing costs [106-108]. For most of medical history, predictions relied primarily on the physician's knowledge and experience based on prior patient histories and medical records. However, studies have indicated that physicians tend to execute poorly in predicting the prognosis and survival expectancy, often over-predicting survival time [109-111]. Statistical algorithms, such as the Cox proportional-hazards model [112], have been implemented to assist physicians' prediction in many studies [113-116], but they are not particularly accurate [12]. As a comparison, ML has shown its potential to predict a patient's prognosis and survival in genomic, transcriptomic, proteomic, radiomic, and other data sets. Chen et al. [117] used 3-year-survival as a threshold to split the patients into high risk (survival time <36 months) and low risk (survival time >36 months) groups, and then constructed a neural network model to binary predict the risk of a patient using his gene expression data and clinical variables. In their model, they tested four microarray gene expression data sets and achieved an overall accuracy of 83.0% with only five identified survival-time correlated genes. Liu et al. [118] also utilized gene expression data for a 3-year-survival classification. Unlike Chen et al. [117], the authors integrated three types of sequencing data -- RNA-sequence, DNA methylation and DNA mutation -- to select a total of 22 genes to promote their model's stability. Meanwhile, LUADpp [119] and Cho et al. [120] used the somatic mutations as input features to model a 3-year-survival risk classification. To select the highest significant mortality-associated-genes, Cho et al. [120] used Chi-squared tests, and LUADpp [119] used a published genome-wide rate comparison test [121] that was able to balance statistical power and precision to compare gene mutation rates. Due to the complexity of survival prediction, multi-omics tumor data have been integrated for analysis in many studies. Compared with single-omics data, the multi-omics data is more challenging to accurately extract the most significant genes for prediction. To address the issue, several studies [122-125] designed a similar workflow. They first constructed a matrix representing the similarity between patients based on their multi-omics data. Using the obtained matrix, they then employed an unsupervised clustering model (usually autoencoder with K-means clustering) to categorize the patients into two clusters. The two clusters were labeled "high-risk" and "low-risk" in terms of the different survival outcomes between the two clusters in the Kaplan–Meier analysis. Following the survival outcome differences, the mortality-associated-genes were extracted using a statistical model [122, 123] or a ML model [124, 125] for downstream analyses.

## Apply ML for lung cancer immunotherapy

### Immunotherapy response prediction

Immunotherapy has become increasingly important in recent years. It enables a patient's own immune system to fight cancer, in most cases, by stimulating T cells. Up to date, distinct novel immunotherapy treatments are being tested for lung cancer, and a variety of them have become standard parts of immunotherapy. Immune checkpoint inhibitors (ICIs), especially programmed cell death protein 1 (*PD-1*)/programmed cell death protein ligand 1 (*PD-L1*) blockade therapy [126], have demonstrated to be valuable in the treatment of patients with non-small cell lung cancer (NSCLC) [127, 128]. However, immunotherapy is not yet as widely used as surgery, chemotherapy, or radiation therapies. One interpretation is that it doesn't work for all patients due to the uniqueness of a patient's tumor immune microenvironment (TIME). Therefore, estimating whether a patient will respond to immunotherapy is important for cancer treatment. Recently, AI-based technologies have been developed to predict immunotherapy responses based on immune genomic signatures and medical imaging signatures [129]. To predict the response to *PD-1/PD-L1* blockade therapy, Wiesweg et al.

[130] utilized gene expression profiles of 770 targeted genes as input features to train four classifiers (SVM, RF, LR and XGBoost) for RECIST classification. Aside from genomic data, features from CT scans can also be used to assess the RECIST level of a patient. Two recent studies [131, 132] used radiomic biomarkers as well as other imaging features of tumor lesions from contrast-enhanced computed tomography (CE-CT) scans to train a classifier, including LR and RF, for RECIST classification.

**Tumor-infiltrating lymphocytes (TILs) evaluation**

The proportion of TILs is another important metric for immunotherapy response evaluation. To this end, using transcriptomics data, DeepTIL [133] optimized the cell deconvolution model CIBERSORT [134] to automatically compute the abundance of the leucocyte subsets (B cells, *CD4*+ T cells, *CD8*+ T cells, γδ T cells, Mo-Ma-DC cells and granulocytes) within a tumor sample. A different approach [135] utilized a total of 84 radiomic features from the CE-CT scans, along with RNA-seq of 20,530 genes as biomarkers to train a linear elastic-net regression model to predict the abundance of *CD8* T-cells. Another study [136] created a deep learning model to identify TILs in digitized H&E-stained images. The methodology consisted of two unique CNN modules to evaluate TILs at different scales: a lymphocyte infiltration classification CNN (lymphocyte CNN) and a necrosis segmentation CNN (necrosis CNN). The "lymphocyte CNN" aimed to categorize the input image into with- and without-lymphocyte infiltration regions. It consists of two steps: a convolutional autoencoder (CAE) [137] for feature extraction, followed by a VGG 16-layer network [138] for TIL region classification. The "necrosis CNN" aimed to detect TILs within a necrosis region. They used the DeconvNet [139] model for TIL segmentation in "necrosis CNN" as the model has been shown to achieve high accuracy with several benchmark imaging datasets.

**Neoantigen prediction**

In addition to immunotherapy response prediction, ML algorithms have shed light on neoantigen prediction for immunotherapy. Neoantigens are tumor-specific mutated peptides generated by somatic mutations in tumor cells, which can induce antitumor immune responses [140-142]. Recent work has demonstrated that immunogenic neoantigens benefit the development and optimization of neoantigen targeted immune therapies [143-146]. In accordance with neoantigen studies in clinical trials, state-of-the-art ML approaches have been implemented to identify neoantigens based on HLA class I and II processing and presentation [147-151]. Using the identified somatic mutations, ML models can estimate the binding affinity of the encoded mutated peptides to the patient's HLA alleles (peptide–HLA binding affinity). The neoantigens can be further predicted based on the estimated peptide–HLA binding affinity. NetMHC [152, 153] utilized a receptor-ligand dataset consisting of 528 peptide–HLA binding interactions measured by Buus et al. [154] to train a combination of several NNs for neo-peptide affinity prediction. To make the prediction more accurate, NetMHC-pan [155, 156] used a larger data set consisting of 37,384 unique peptide-HLA interactions covering 24 HLA-A alleles and 18 HLA-B alleles (26503 and 10881 for the A and B alleles, respectively) to train their NN model. Both tools have been implemented to study the neoantigen landscape in lung cancers [140, 157-159].

## Challenges and future perspectives

This review depicts the applications of ML algorithms in lung cancer early detection, diagnosis decision, prognosis prediction, drug response evaluation, and immunotherapy practice (Table S1 and S2). Despite the widespread use of ML studies in lung cancer clinical practice and research, there are still challenges to be addressed. Here, we post four major challenges and perspectives for future studies.

### Imaging data analysis

Learning how to effectively extract nuance from imaging data is critical for clinical use. In the earlier ML-based CAD system, feature extractions were typically based on the image intensity, shape, and texture of a suspicious region along with other clinical variables [160]. However, these approaches are arbitrarily-defined and may not retrieve the intrinsic features of a suspicious nodule. To this end, a CNN-based CAD system was developed leveraging CNN models to extract features directly from raw imaging data with multilevel representations and hierarchical abstraction [161-163]. Contrary to previous methods, features from a CNN model are not designed by humans, and reflect the intrinsic features of the nodule in an objective and comprehensive manner. Recently, the Vision Transformer (ViT) has emerged as the current state-of-the-art in computer vision [164, 165]. In comparison to CNN, ViT outperformed almost ×4 in terms of computational efficiency and accuracy, and was more robust when training on smaller datasets [166]. Although, to our knowledge, ViT models haven't been implemented in any lung cancer imaging studies, they have shown their potential as a competitive alternative to CNN in imaging data analysis.

### Multi-omics data integration and analysis

Though the ITH in cancer causing drug resistance challenges our ability to characterize tumors [167], multi-omics data provides a comprehensive insight into the molecular functions of lung cancer studies. However, large multi-omics data sets, especially the recent development of single-cell-based [168] and spatial-based [169] technologies, leading to computationally intensive analysis is a major challenge [170]. Multi-omics analysis can be broken down into three steps: processing, integration and analysis. Most datasets are sequenced from different platforms, thus sequencing bias and background noise inevitably exist within these platforms, making the first and second steps difficult to address perfectly. Removing batch effects and putting datasets from multiple platforms together in a framework that allows us to further analyze the mechanisms of cancer drug resistance and recurrence is important for cancer therapies. Though biomedical studies have experimented and/or benchmarked integrative tools [170-173], they are not comprehensive and discriminating enough to address the choice of tools in the context of biological questions of interest.

### Immunotherapy

ML has shown its capacity for personalized immunotherapy approaches and provides guidance on the combination of immunotherapy with other treatments for lung cancer patients. However, unlike chemotherapy or surgery that have abundant clinical trials, the clinical trials of immunotherapy are limited, and the available patients for a trial are usually insufficient for some ML models requiring large amounts of training data [129, 174]. Therefore, integrating data sets from different clinical trials and reducing overfitting in small samples is necessary to reinforce a model's performance in immunotherapy practices. Despite these improvements, most patients fail ICI therapy due to drug resistance or non-responsive [175]. Thus, identifying neoantigens is valuable for immunotherapy studies. Although ML models have been proposed to predict HLA binding, a limited number of neoantigens have been approved for clinical trials. Further study of neoantigen prediction requires both efforts in ML model design and clinical practice.

**Clinical decision making**

A recent study estimated that the overall costs for lung cancer therapy would exceed $50,000 [176] for most patients, and that the cost would be high for most families. Thus, using ML in predicting the effectiveness of a therapy and optimizing the combination of different therapies will pave the way for personalized treatment. However, most existing ML models for clinical decision making have difficulty in keeping up with knowledge evolution and/or dynamic health care data change [177]. Currently, clinical decision-support systems, including IBM Watson Health and Google DeepMind Health, have been implemented in lung cancer treatments in recent years [178, 179]. Although the efficiency of clinical work has improved with the help of these systems, they are still far from perfect in terms of clinical trials, and currently cannot replace physicians at this stage [179]. We still have a long way to go before we realize the full potential of ML clinical decision making tools.

## Authors' contributions

YLuo and YLi conceived and designed the research; YLi and XW collected the data; YLi and XW contributed to the figure; YLuo, YLi and XW drafted the manuscript; PY and GJ provided critical revision. All authors have read, edited and approved the final manuscript.

## Competing interests

The authors have declared no competing interests.

## Acknowledgements

This study is supported in part by US National Institutes of Health (Grant No. U01TR003528 and R01LM013337).

# Figures

## Figure 1 Applications of machine learning model in lung cancer

Abbreviations: CT: computed tomography; MALDI: matrix-assisted laser desorption/ionization; CNN: convolutional neural network; cfDNA: cell free DNA; CAD: computer-aided diagnosis; CNV: copy number variation; RECIST: response evaluation criteria in solid tumors; TIL: tumor-infiltrating lymphocytes

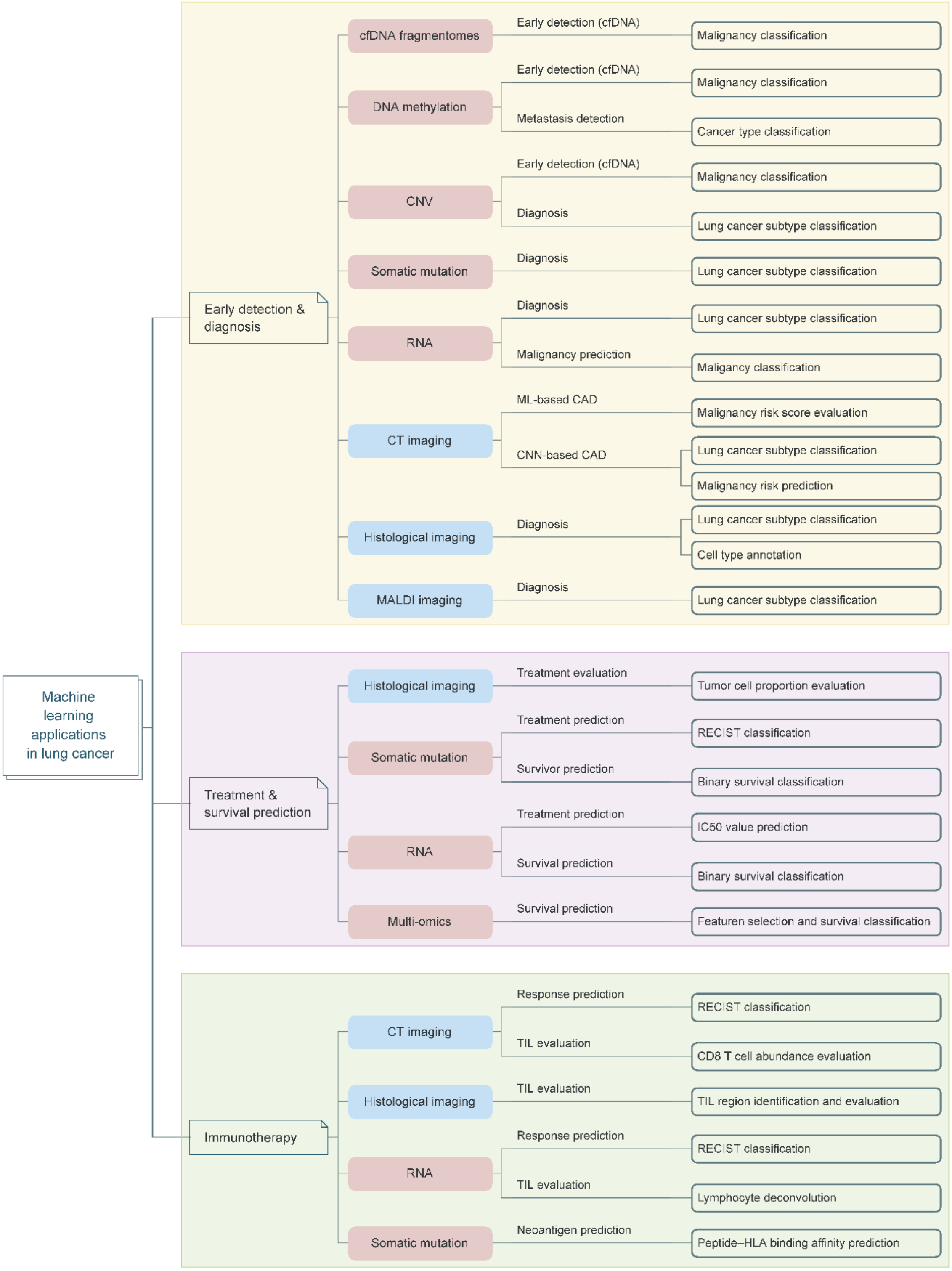

**Tables**

**Table 1 Publications relevant to ML on early detection and diagnosis using sequencing.**

| Name | ML methods | NO. of samples | Sequencing data type | Performance | Validation method | Feature selection |
|---|---|---|---|---|---|---|
| Mathios et al. [75] | LR model with a LASSO penalty | 799 | cfDNA fragment | AUC = 0.98 | 10-fold cross-validation | cfDNA fragment feature |
| Lung-CLiP [76] | 5-nearest neighbor; 3-nearest neighbor; NB; LR; DT | 160 | cfDNA | AUC from 0.69 to 0.98 | Leave-one-out cross validation | SNV + CNV features |
| Liang et al. [77] | LR | 296 | ctDNA | AUC = 0.816 | 10-fold cross-validation | Nine DNA methylation markers |
| Kobayashi et al. [80] | Diet Networks with EIS | 954 | somatic mutation | Accuracy = 0.8 | 5-fold cross-validation | SNVs, insertions, and deletions across 17,961 unique gene symbols |
| Whitney et al. [83] | LR | 299 | RNA-seq of BECs | AUC = 0.81 | 10-fold cross-validation | RNA of clinical covariates (gender, tobacco use, and smoking history) associated genes + RNA of lung cancer-associated genes |
| Podolsky et al. [84] | KNN; NB normal distribution of attributes; NB distribution through histograms; SVM; C4.5 DT | 529 | RNA-seq | AUC = 0.91 | Hold-out | RNA-seq |
| Choi et al. [85] | An ensemble model based on elastic net LR, SVM, hierarchical LR | 2,285 | RNA-seq of bronchial brushing sample | AUC = 0.74 | 5-fold cross-validation | RNA-seq of 1,232 genes and four clinical covariates (age, pack-years,

| | | | | | | |
|---|---|---|---|---|---|---|
| | | | | | | inhaled medication use, specimen collection timing) |
| Aliferis et al. [86] | linear SVM; polynomial-kernel SVM; KNN; NN | 203 | RNA-seq | AUC from 0.8783 to 0.9980 | 5-fold cross-validation | RNA-seq of selected genes using RFE and UAF |
| Aliferis et al. [88] | DT; KNN; linear SVM; polynomial-kernel SVM; RBF-kernel SVM; NN | 37 | Gene copy number measure by array CGH | Accuracy = 0.892 | Leave-one-out cross validation | Gene copy number of 80 best genes according to weights in linear SVM trained with all genes |
| Raman et al. [78] | RF; SVM; LR with ridge, elastic net; LASSO regularization | 843 | cfDNA | mAUC from 0.896 to 0.936 | Leave-one-out cross-validation | Copy number profiling of cfDNA |
| Daemen et al. [89] | LS-SVM | 89 | CNV measured by CGH | Accuracy from 0.880 to 0.955 | 10-fold cross-validation | CNV measured by CGH |
| Jurmeister et al. [90] | NN, SVM, RF | 972 | DNA methylation | Accuracy from 0.878 to 0.964 | 5-fold cross-validation | Top 2000 variable CpG sites |

*Note:* Abbreviations: ctDNA: circulating tumor DNA; cfDNA: cell-free DNA; LR: logistic regression; AUC: area under the curve; mAUC: Mean area under the curve; EIS: element-wise input scaling; NB: naive Bayes; DT: decision tree; CNV: copy number variation; SNV: single-nucleotide variant; SVM: support vector machines; BEC: bronchial epithelial cell; CGH: Comparative Genomic Hybridization; LS-SVM: weighted least squares support vector machines; RF: random forest; RFE: recursive feature elimination; UAF: univariate association filtering; NN: neural network.

# Supplementary material

## Table S1 Lung cancer benchmark datasets used by the machine learning methods reviewed in this paper

| NO. | Database | Data type | Website |
|---|---|---|---|
| 1 | TCGA [5] | Genomics data | https://www.cancer.gov/about-nci/organization/ccg/research/structural-genomics/tcga |
| 2 | TCIA [6] | Image (CT, MRI, PET, etc.) | https://www.cancerimagingarchive.net/ |
| 3 | PanCan [180] | CT | https://www.thelancet.com/journals/lanonc/article/PIIS1470-2045(17)30597-1/fulltext |
| 4 | BCCA [181] | CT | https://www.atsjournals.org/doi/full/10.1164/rccm.200301-144OC |
| 5 | DLCST [182] | CT | https://www.sciencedirect.com/science/article/pii/S1556086415316786 |
| 6 | Kriegsmann et al. [34] | MALDI | https://www.ncbi.nlm.nih.gov/pmc/articles/PMC5054336/ |
| 7 | LIDC-IDRI [183] | CT | https://wiki.cancerimagingarchive.net/display/Public/LIDC-IDRI |
| 8 | LTRC | CT | https://ltrcpublic.com/ |
| 9 | NLST | CT | https://www.cancer.gov/types/lung/research/nlst |
| 10 | NELSON [184] | CT | https://acsjournals.onlinelibrary.wiley.com/doi/10.1002/cncr.23590 |
| 11 | Venkadesh et al. [45] | CT | https://pubs.rsna.org/doi/full/10.1148/radiol.2021204433 |
| 12 | Jiang et al. [99] | CT | https://ieeexplore.ieee.org/document/8417454 |
| 13 | TRACERx [185] | Histological image | https://www.nejm.org/doi/10.1056/NEJMoa1616288?url_ver=Z39.88-2003&rfr_id=ori:rid:crossref.org&rfr_dat=cr_pub%20%200www.ncbi.nlm.nih.gov |
| 14 | TCGA – LUSC [186] | Histological image; Genomics | https://www.nature.com/articles/nature11404 |
| 15 | TCGA – LUAD [187] | Histological image; Genomics | https://www.nature.com/articles/nature13385 |
| 16 | Pan-Lung Cancer dataset [188] | Exome sequences and copy number profiles | https://www.nature.com/articles/ng.3564#Sec20 |
| 17 | AEGIS [189] | RNA | https://www.ncbi.nlm.nih.gov/geo/query/acc.cgi?acc=GSE4115 |
| 18 | Bhattacharjee et al. [190] | RNA | https://www.pnas.org/content/98/24/13790.long |
| 19 | Beer et al. [191] | RNA | https://www.nature.com/articles/nm733#Sec3 |
| 20 | Wigle et al. [192] | RNA | https://cancerres.aacrjournals.org/content/62/11/3005.long |
| 21 | Gordon et al. [193] | RNA | https://cancerres.aacrjournals.org/content/62/17/4963.long |
| 22 | Silvestri et al. [194] | RNA | https://www.nejm.org/doi/full/10.1056/NEJMoa1504601 |
| 23 | Aliferis et al. [88] | Array CGH | https://www.ncbi.nlm.nih.gov/pmc/articles/PMC2244172/pdf/procamiasymp00001-0048.pdf |
| 24 | Raman et al. [78] | Cell-free DNA | https://genomemedicine.biomedcentral.com/articles/10.1186/s13073-020-00735-4#Sec2 |
| 25 | CLCGP project [195] | Genomics data | https://www.science.org/doi/10.1126/scitranslmed.3006802 |
| 26 | Daemen et al. [89] | Array CGH | http://psb.stanford.edu/psb-online/proceedings/psb09/daemen.pdf |
| 27 | NCI caArray database [196] | RNA | https://wiki.nci.nih.gov/display/caArray2/caArray+Retirement+Announcement |
| 28 | Wang et al. [197] | Somatic mutation in EGFR | https://www.nature.com/articles/srep02855#Sec9 |
| 29 | Lee et al. [198] | Genomic DNA in EGFR | https://www.sciencedirect.com/science/article/pii/S1556086415334705?via%3Dihub#bib18 |
| 30 | EGFR Mutation Database [199] | Somatic mutation in EGFR | http://www.cityofhope.org/cmdl/egfr_db |
| 31 | Zou et al. [200] | Somatic mutation in EGFR | https://www.nature.com/articles/s41598-017-06632-y#Sec11 |
| 32 | Garnett et al. [201] | RNA | https://www.nature.com/articles/nature11005#ethics |
| 33 | Wiesweg et al. [130] | RNA | https://www.sciencedirect.com/science/article/pii/S0959804920305219?via%3Dihub |
| 34 | Trebeschi et al. [131] | CT | https://www.sciencedirect.com/science/article/pii/S0923753419312025?via%3Dihub |
| 35 | Coroller et al. [132] | CT | https://www.sciencedirect.com/science/article/pii/S0167814016310386?via%3Dihub |
| 36 | GEO | RNA | https://www.ncbi.nlm.nih.gov/gds |
| 37 | MOSCATO database [202] | CT, RNA | https://cancerdiscovery.aacrjournals.org/content/7/6/586.long#sec-8 |
| 38 | Champiat et al. [203] | CT | https://clincancerres.aacrjournals.org/content/23/8/1920.long#sec-6 |
| 39 | Sun et al. [204] | CT | https://www.sciencedirect.com/science/article/pii/S095980491731153X |
| 40 | Buus et al. [154] | Tumor peptidomics dataset | https://www.sciencedirect.com/science/article/pii/030441659400172T?via%3Dihub |
| 41 | Peters et al. [205] | Tumor peptidomics dataset | https://journals.plos.org/ploscompbiol/article?id=10.1371/journal.pcbi.0020065 |
| 42 | Bulik-Sullivan et al. [151] | Tumor peptidomics dataset | https://massive.ucsd.edu/ProteoSAFe/dataset.jsp?task=ad676ada8227478e92996c2ef849ea31 |
| 43 | Mathios et al. [75] | cfDNA | https://ega-archive.org/studies/EGAS00001005340 |
| 44 | Chabon et al. [76] | cfDNA | https://clip.stanford.edu/ |
| 45 | Liang et al. [77] | DNA methylation | https://www.thno.org/v09p2056.htm |
| 46 | Jurmeister et al. [90] | DNA methylation | https://www.ncbi.nlm.nih.gov/geo/query/acc.cgi?acc=GSE124052 |

*Note:* Abbreviations: TCGA: The Cancer Genome Atlas; TCIA: The Cancer Imaging Archive; PanCan: Pan-Canadian Early Detection of Lung Cancer Study; BCCA: British Columbia Cancer Agency study; DLCST: The Danish Lung Cancer Screening Trial; LIDC-IDRI: Lung Image Database Consortium image collection; MALDI: ML-based CAD Matrix-assisted laser desorption/ionization; LTRC: Lung Tissue Research Consortium; NLST: National Lung Screening Trial; NELSON: Dutch-Belgian randomized lung cancer screening trial; MOSCATO: The Molecular Screening for Cancer Treatment Optimization: CGH: comparative genomic hybridization; cfDNA: cell-free DNA.

## Table S2 Machine learning methods used for benchmark studies in lung cancer therapy

| Name | Application scenarios | Datasets in Table S1 | Benchmarks |
|---|---|---|---|
| McWilliams et al. [31] | ML on early detection and diagnosis using medical imaging datasets | 3, 4 | No |
| Riel et al. [32] | ML on early detection and diagnosis using medical imaging datasets | 5 | Radiologists |
| Wille et al. [33] | ML on early detection and diagnosis using medical imaging datasets | 3, 4, 5 | No |
| Kriegsmann et al. [34] | ML on early detection and diagnosis using medical imaging datasets | 6 | No |
| Hussein et al. [38] | ML on early detection and diagnosis using medical imaging datasets | 7 | GIST features [206] with LASSO; 3D CNN (Karpathy et al. [207]) multi-task learning with trace norm |
| Khosravan et al. [39] | ML on early detection and diagnosis using medical imaging datasets | 8 | Khosravan et al. [208]; Dou et al. [209]; Radiologists |
| Ardila et al. [48] | ML on early detection and diagnosis using medical imaging datasets | 7, 9 | No |
| Ciompi et al. [40] | ML on early detection and diagnosis using medical imaging datasets | 10 | RF; SVM; Radiologists |
| Buty et al. [43] | ML on early detection and diagnosis using medical imaging datasets | 7 | No |
| Venkadesh et al. [45] | ML on early detection and diagnosis using medical imaging datasets | 11 | Radiologists; PanCan model [180] |
| AbdulJabbar et al. [53] | ML on early detection and diagnosis using medical imaging datasets | 13 | No |
| Coudray et al. [56] | ML on early detection and diagnosis using medical imaging datasets | 14, 15 | No |
| Mathios et al. [75] | ML on early detection and diagnosis using -omics sequencing datasets | 1, 43 | No |
| Lung-CLiP [76] | ML on early detection and diagnosis using -omics sequencing datasets | 44 | 5-nearest neighbor; 3-nearest neighbor; NB; LR; DT |
| Liang et al. [77] | ML on early detection and diagnosis using -omics sequencing datasets | 45 | No |
| Kobayashi et al. [80] | ML on early detection and diagnosis using -omics sequencing datasets | 16 | MLP; Diet Networks [210] |
| Whitney et al. [83] | ML on early detection and diagnosis using -omics sequencing datasets | 17 | No |
| Podolsky et al. [84] | ML on early detection and diagnosis using -omics sequencing datasets | 18, 19, 20,21 | KNN; NB; SVM; DT |
| Choi et al. [85] | ML on early detection and diagnosis using -omics sequencing datasets | 22 | RF; SVM; LDA; GB; penalized LR |
| Aliferis et al. [86] | ML on early detection and diagnosis using -omics sequencing datasets | 18 | linear SVM; polynomial-kernel SVM; KNN; NN |
| Aliferis et al. [88] | ML on early detection and diagnosis using -omics sequencing datasets | 23 | DT; KNN; linear SVM; polynomial-kernel SVM; RBF-kernel SVM; NN |
| Raman et al. [78] | ML on early detection and diagnosis using -omics sequencing datasets | 24, 25 | RF; SVM; LR with ridge; Elastic Net [211]; Lasso regularization [212] |
| Daemen et al. [89] | ML on early detection and diagnosis using -omics sequencing datasets | 26 | No |
| Jurmeister et al. [90] | ML on early detection and diagnosis using -omics sequencing datasets | 46 | NN, SVM, RF |
| Chen et al. [117] | Survival prediction | 27 | No |
| LUADpp [119] | Survival prediction | 1 | No |
| Cho et al. [120] | Survival prediction | 1 | NB; KNN; SVM; DT |
| Yu et al. [122] | Survival prediction | 1 | No |
| CIMLR [123] | Survival prediction | 1 | iCluster+ [213]; Bayesian consensus clustering [214]; PINS [215]; SNF [216] |
| Takahashi et al. [125] | Survival prediction | 1 | No |
| Asada et al. [124] | Survival prediction | 1 | SVM; KNN; RF; LR |
| Qureshi [101] | Prognosis and drug response prediction | 28, 29, 30, 31 | Wang et al. [197]; Ma et al. [217]; Duan et al. [218]; Zou et al. [219]; Kureshi et al. [220] |
| Kapil et al. [102] | Prognosis and drug response prediction | Not publicly available | No |
| Jiang et al. [99] | Prognosis and drug response prediction | 2, 7, 12 | FRRN [100]; Unet [221]; SegNet [222]; RF+fCRF [223] |
| Geeleher et al. [104] | Prognosis and drug response prediction | 32 | RF; PAM [224]; Principal component regression [225]; Lasso regression [212]; Elastic Net regression [211] |
| Liu et al. [118] | Prognosis and drug response prediction | 1, 36 | SVM; RF; LR: NB; linear regression; SVR (kernel Poly); SVR (kernel Linear); ridge regression |
| Wiesweg et al. [130] | Immunotherapy response prediction | 33 | No |
| Trebeschi et al. [131] | Immunotherapy response prediction | 34 | No |
| Coroller et al. [132] | Immunotherapy response prediction | 35 | No |
| DeepTIL [133] | Tumor-infiltrating lymphocytes (TILs) evaluation | 36 | CIBERSORT [134] |
| Sun et al. [135] | Tumor-infiltrating lymphocytes (TILs) evaluation | 1, 2, 37, 38, 39 | No |
| Saltz et al. [136] | Tumor-infiltrating lymphocytes (TILs) evaluation | 1 | Zhao et al. [226] |
| NetMHC [152] | Neoantigen prediction | 40 | No |
| NetMHC-pan [156] | Neoantigen prediction | 41 | NetMHC [152] |
| Bulik-Sullivan et al. [151] | Neoantigen prediction | 42 | NetMHC [152]; MHCflurry [227]; NetMHCpan [156] |

*Note:* Abbreviations: CNN: convolutional neural network; KNN: k-nearest neighbors; NB: naive Bayes; RF: random forests; SVM: support vector machine; SVR: support vector regression; LR: logistic regression; DT: decision tree; NN: neural network; LDA: linear discriminant analysis; GB: gradient boosting; MLP: multilayer perceptron; RF+fCRF: Random forest with fully connected conditional random field